\documentclass[aps,pre,preprint,superscriptaddress,longbibliography]{revtex4-2}

\usepackage[utf8]{inputenc}
\usepackage[T1]{fontenc}
\usepackage{amsmath, amssymb, amsfonts}
\usepackage{bm}
\usepackage{graphicx}
\usepackage{booktabs}
\usepackage{multirow}
\usepackage{xcolor}
\usepackage[colorlinks=true,linkcolor=blue,citecolor=blue,urlcolor=blue]{hyperref}

\newcommand{\rmse}{\mathrm{RMSE}}

\begin{document}

\title{Potential-energy gating for robust state estimation
in bistable stochastic systems}

\author{Luigi Simeone}
\affiliation{Independent Researcher, Turin, Italy}
\email{luigi.simeone@tech-management.net}

\date{\today}

\begin{abstract}
We introduce potential-energy gating, a method for robust state estimation
in systems governed by double-well stochastic dynamics.
The core idea is to modulate the observation noise covariance of a Bayesian
filter using the local value of a known or assumed potential energy function:
observations are trusted when the state sits near a potential minimum and
progressively discounted as the state approaches the barrier separating
metastable wells.
This physics-based mechanism differs from purely statistical robust filters,
which treat all regions of state space identically, and from constrained
filters, which impose hard bounds on the state rather than modulating
observation trust.
The approach is particularly relevant for non-ergodic or data-scarce domains
where only a single realization is available and purely statistical methods
cannot reliably learn the noise structure from data alone.
We implement the gating within Extended, Unscented, Ensemble, and Adaptive
Kalman filters as well as particle filters, requiring only two additional
hyperparameters beyond the standard formulation.
Synthetic benchmarks on a Ginzburg-Landau double-well process with $10\%$
outlier contamination and Monte Carlo validation over $100$ independent
replications show that potential-energy gating improves root-mean-square
estimation error by $57$--$80\%$ relative to the standard Extended Kalman
Filter, with all improvements statistically significant
($p < 10^{-15}$, Wilcoxon signed-rank test).
A naive topological baseline that uses only the distance to the nearest
well---without the potential energy profile---achieves $57\%$ improvement,
confirming that the continuous energy landscape contributes an additional
$\sim\!21$ percentage points.
The method is robust to parameter misspecification: even when the assumed
potential parameters deviate by $50\%$ from their true values, the
improvement never falls below $47\%$.
A supplementary comparison between externally forced and spontaneous
(Kramers-type) transitions shows that the gating mechanism retains $68\%$
improvement under noise-induced transitions, whereas the naive baseline
degrades to $30\%$.
As an empirical illustration, we apply the framework to Dansgaard-Oeschger
events in the NGRIP $\delta^{18}$O ice-core record, estimating an asymmetry
parameter $\gamma = -0.109$ (bootstrap $95\%$ CI: $[-0.220, -0.011]$,
excluding zero) and demonstrating that outlier fraction explains $91\%$
of the variance in filter improvement.
\end{abstract}

\maketitle

\section{Introduction}
\label{sec:intro}

Bistable stochastic systems---processes whose dynamics are governed by a
double-well potential energy landscape---arise across the physical,
biological, and social sciences~\cite{Haken1983}.
Examples include Josephson junctions in condensed matter
physics~\cite{Kramers1940}, gene regulatory switches in molecular
biology~\cite{Gardiner2009}, abrupt climate transitions in
paleoclimatology~\cite{Ditlevsen1999}, and regime-switching dynamics
in quantitative finance~\cite{Hamilton1989}.
In all these systems, the coexistence of two metastable states separated
by an energy barrier creates a fundamental challenge for state estimation:
during transitions between wells, observations become unreliable because
the system traverses a region of high potential energy where the restoring
force vanishes and noise dominates the dynamics.

Standard Bayesian filtering---the Extended Kalman Filter (EKF) and its
variants~\cite{Kalman1960,Sarkkae2013}---treats observation noise as
homogeneous across state space.
When the observation channel is contaminated by outliers (measurement errors
whose magnitude exceeds typical process fluctuations), this assumption leads
to catastrophic state estimate corruption, because the filter cannot
distinguish a legitimate transition between wells from an anomalous
measurement.
Existing approaches to outlier-robust filtering fall into three categories,
none of which exploits the physics of the energy landscape.

First, \emph{statistical gating} methods, such as chi-square gating based on
the Mahalanobis distance~\cite{BarShalom2001,Chang2014}, reject observations
that are statistically surprising according to the filter's own innovation
covariance.
This mechanism is binary (accept or reject), physics-agnostic, and fails
precisely when it is needed most: during genuine transitions, the innovation
is large \emph{because} the state is changing, so statistical gating may
reject valid measurements.

Second, \emph{heavy-tailed filters} replace Gaussian noise assumptions with
Student-$t$ distributions~\cite{Agamennoni2012,Roth2013,Roth2017}, providing
soft down-weighting of outlying observations.
While effective for uniformly contaminated data, these methods apply the same
degree of robustness everywhere in state space, with no mechanism to exploit
the fact that observations near well minima are physically more reliable than
observations near the barrier.

Third, \emph{regime-switching models}~\cite{Hamilton1989} and the Interacting
Multiple Model (IMM) algorithm~\cite{Blom1988} assign discrete labels to
regimes and switch between parameter sets.
These approaches impose a discrete, finite number of states rather than a
continuous energy landscape, and the switching probabilities are typically
estimated from data rather than derived from physics.

The need for physics-based filtering is especially acute in domains where
only a single realization of the process is available---paleoclimate records,
seismological catalogs, or financial time series of rare regime changes.
In such non-ergodic or data-scarce settings, purely statistical methods
cannot reliably learn the noise structure from the data alone, because the
number of observed transitions is too small for robust parameter estimation.
Incorporating prior knowledge about the energy landscape offers a principled
alternative: the physics constrains the filter where the data cannot.

The present work proposes a fourth approach: \emph{potential-energy gating},
in which the observation noise covariance $R$ is modulated by the local value
of a known or assumed potential energy function $V(x)$.
When the state estimate $\hat{x}$ sits near a potential minimum
($V(\hat{x}) \approx 0$), observations are trusted at their nominal
uncertainty $R_0$.
As $\hat{x}$ moves toward the barrier ($V(\hat{x}) \to V_{\max}$), the
effective observation noise is inflated:
\begin{equation}
\label{eq:gating_preview}
R_{\mathrm{eff}}(x) = R_0 \cdot \bigl[1 + g\, V(x)\bigr],
\end{equation}
where $g > 0$ is a gating sensitivity parameter.
This mechanism is continuous, state-dependent, and physics-based: it encodes
the physical intuition that observations of a system traversing an unstable
equilibrium carry less information about the underlying state than
observations of a system resting in a metastable well.

\subsection{Relation to prior work}
\label{sec:prior}

The use of double-well potentials for modeling abrupt climate transitions has
a rich history.
Ditlevsen~\cite{Ditlevsen1999} established the paradigm of
Dansgaard-Oeschger (D-O) events as noise-induced transitions in a bistable
potential, fitting a Langevin model to the GRIP Ca$^{2+}$ record.
Kwasniok and Lohmann~\cite{Kwasniok2009} used the Unscented Kalman Filter
(UKF) for maximum-likelihood estimation of double-well SDE parameters from
the NGRIP $\delta^{18}$O record, while Kwasniok~\cite{Kwasniok2012} extended
this framework to disentangle dynamical and observational noise using
predictive likelihood maximization.
Livina, Kwasniok, and Lenton~\cite{Livina2010} developed potential
analysis---fitting polynomial potentials to empirical probability densities---to
detect changes in the number of climate states over the last 60\,kyr.
Livina and Lenton~\cite{Livina2007} proposed bifurcation detection via
detrended fluctuation analysis as an early-warning signal.
Lohmann and Ditlevsen~\cite{Lohmann2019} used Approximate Bayesian
Computation to compare double-well and relaxation oscillator models, finding
evidence favoring the oscillator for D-O dynamics.

All of these contributions address the \emph{inverse problem}: reconstructing
the potential from data.
Our work addresses a different and complementary question---the \emph{forward
problem} of using a known or assumed potential to improve real-time state
estimation.
The Kalman filter in Kwasniok's work serves as a parameter estimation engine;
in our framework, the filter is the end product, and the potential serves as
prior physical knowledge that modulates its observation model.

In the broader context of physics-informed inference, constrained Kalman
filtering~\cite{Simon2010} imposes hard bounds on the state vector but does
not modify the observation model.
Physics-informed neural networks (PINNs)~\cite{Raissi2019,Karniadakis2021}
embed PDE residuals in training loss functions but operate in batch mode
rather than sequential filtering.
Data-driven methods such as SINDy~\cite{Brunton2016} discover governing
equations from data but require high-frequency, low-noise sampling.
Adaptive methods for jointly estimating model and observation error
covariances~\cite{Tandeo2020} infer $R$ from the statistical properties of
innovation sequences rather than prescribing it from physics.
The present approach is, to our knowledge, the first to embed physics in the
\emph{observation reliability channel} of a Bayesian filter: modulating $R$
rather than constraining $x$ or regularizing the process model.

\subsection{Scope and organization}

This paper is methodological in scope.
We propose and validate a general mechanism---potential-energy gating---for
any system whose dynamics can be approximated by a double-well potential.
The NGRIP paleoclimatic application in Sec.~\ref{sec:ngrip} serves as an
empirical illustration of how the method can be applied to real data, not as a
claim about the optimal model for D-O events.
We explicitly acknowledge the work of Lohmann and
Ditlevsen~\cite{Lohmann2019}, which has shown that a relaxation oscillator
model may outperform the simple double-well for D-O dynamics; our use of the
Ginzburg-Landau potential for NGRIP is intended to demonstrate the gating
mechanism on a well-known dataset, not to argue that the quartic potential is
the best physical model for this system.

The paper is organized as follows.
Section~\ref{sec:model} defines the state-space model and the
Ginzburg-Landau potential.
Section~\ref{sec:method} derives the potential-gated filter.
Section~\ref{sec:synthetic} presents synthetic validation, including Monte
Carlo benchmarks, misspecification analysis, and the naive topological
baseline.
Section~\ref{sec:ngrip} applies the framework to NGRIP data.
Section~\ref{sec:discussion} discusses limitations and
Sec.~\ref{sec:conclusion} concludes.

\section{State-space model and potential energy}
\label{sec:model}

\subsection{Continuous-time dynamics}

We consider a scalar state $x(t) \in \mathbb{R}$ evolving according to the
overdamped Langevin equation
\begin{equation}
\label{eq:langevin}
\mathrm{d}x = -V'(x)\,\mathrm{d}t + \sigma\,\mathrm{d}W,
\end{equation}
where $V(x)$ is a confining potential, $\sigma > 0$ is the noise intensity,
and $W(t)$ is a standard Wiener process.
The drift term $-V'(x)$ drives the state toward local minima of $V$; the
noise term $\sigma\,\mathrm{d}W$ drives fluctuations and, over sufficiently
long times, transitions between wells.

\subsection{The Ginzburg-Landau potential}

We parametrize $V(x)$ using the Ginzburg-Landau (GL)
form~\cite{GinzburgLandau1950}, the minimal polynomial potential that
exhibits bistability.
In its symmetric version [Eq.~(\ref{eq:GL_symmetric})],
\begin{equation}
\label{eq:GL_symmetric}
V(x) = -\frac{\alpha}{2}\,x^2 + \frac{\beta}{4}\,x^4,
\qquad \alpha, \beta > 0.
\end{equation}
This potential has two degenerate minima at
$x_{\pm} = \pm\sqrt{\alpha/\beta}$, separated by a barrier of height
$\Delta V = \alpha^2/(4\beta)$ located at $x=0$.
We also consider the asymmetric extension
\begin{equation}
\label{eq:GL_asymmetric}
V_\gamma(x) = V(x) - \gamma\,x,
\end{equation}
where $\gamma \in \mathbb{R}$ tilts the potential, breaking the degeneracy
between the two wells.
This form has been used to model the asymmetry between stadial and
interstadial states in glacial climate~\cite{Ditlevsen1999,Kwasniok2009}.

\begin{figure}[tb]
\centering
\includegraphics[width=\columnwidth]{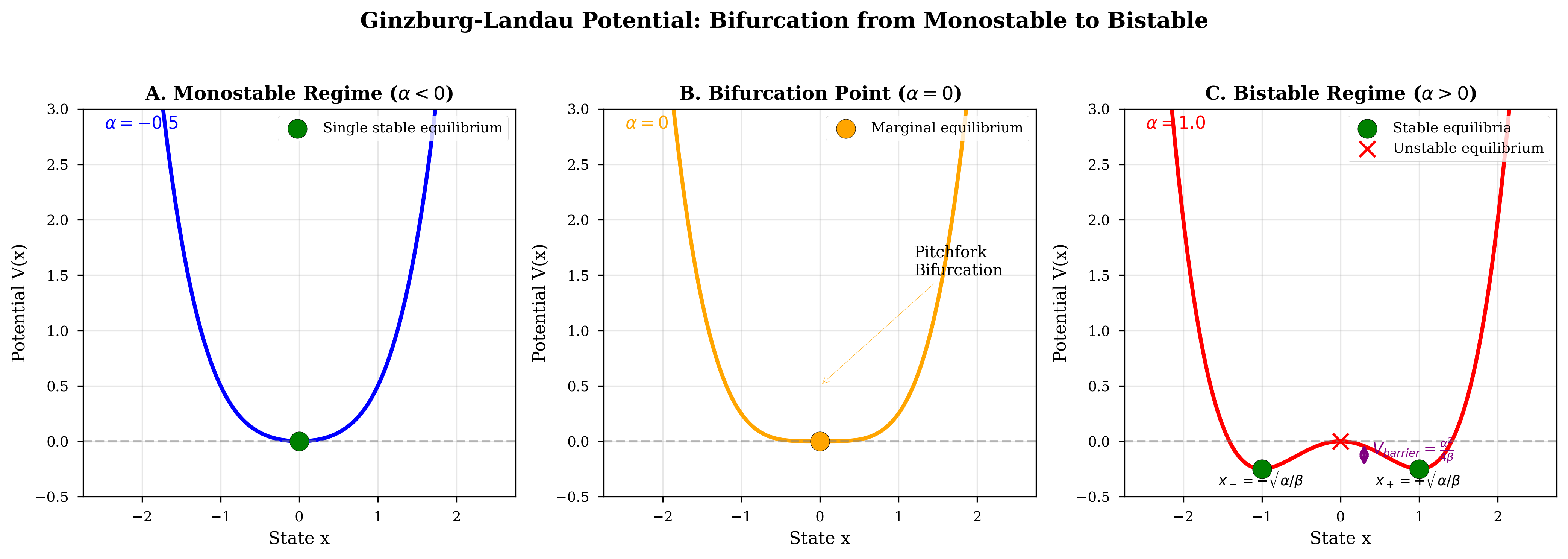}
\caption{%
The Ginzburg-Landau potential $V(x)$ for different parameter regimes.
The symmetric case ($\gamma=0$) exhibits two degenerate wells at
$x_{\pm} = \pm\sqrt{\alpha/\beta}$ separated by a barrier of height
$\Delta V = \alpha^2/(4\beta)$.
The asymmetric case ($\gamma \neq 0$) tilts the landscape, making one well
deeper than the other.
In the potential-energy gating framework, observations are trusted near
the well minima (low $V$) and discounted near the barrier (high $V$).
}
\label{fig:potential}
\end{figure}

Figure~\ref{fig:potential} illustrates the symmetric and asymmetric GL
potential, highlighting the wells, the barrier, and the regions where the
gating mechanism modulates observation trust.

The key physical quantities are: the barrier height $\Delta V$, which sets
the energy scale for transitions; the effective temperature
$T_{\mathrm{eff}} = \sigma^2/2$, which characterizes the thermal fluctuation
energy; and their ratio $\Pi_1 = \Delta V / T_{\mathrm{eff}}$, which governs
the Kramers escape rate~\cite{Kramers1940}.

\subsection{Discrete-time state-space model}

Discretizing Eq.~(\ref{eq:langevin}) with time step $\Delta t$ via the
Euler-Maruyama scheme and adding an observation equation, we obtain the
state-space model
\begin{align}
x_{k+1} &= x_k + f(x_k)\,\Delta t + \sigma\sqrt{\Delta t}\;\eta_k,
\label{eq:process} \\
y_k &= x_k + \varepsilon_k,
\label{eq:observation}
\end{align}
where $f(x) = -V'(x) = \alpha x - \beta x^3 + \gamma$ is the drift,
$\eta_k \sim \mathcal{N}(0,1)$ is the process noise, and $\varepsilon_k$ is
the observation noise [Eq.~(\ref{eq:observation})].
In practice, $\varepsilon_k$ may be heavy-tailed due to measurement errors,
instrument failures, or external contamination.
We model this as a mixture: with probability $(1-p)$,
$\varepsilon_k \sim \mathcal{N}(0, R_0)$ (clean observation), and with
probability $p$, $\varepsilon_k \sim \mathcal{N}(0, R_0 + R_{\mathrm{out}})$
(outlier), where $R_{\mathrm{out}} \gg R_0$.

\section{Potential-gated Bayesian filtering}
\label{sec:method}

\subsection{Motivation: energy-dependent observation reliability}
\label{sec:motivation}

The central observation motivating our approach is that in a bistable system,
the \emph{physical reliability} of an observation depends on where the system
sits in the energy landscape.
Near a potential minimum, the restoring force $|f(x)|$ is large, fluctuations
are bounded, and an observation provides high-quality information about the
latent state.
Near the barrier top, the restoring force vanishes, the system is maximally
sensitive to perturbations, and an observation carries less information
because it cannot distinguish the true state from noise excursions or
incipient transitions.

This motivates replacing the constant observation noise covariance $R_0$ with
a state-dependent covariance, as previewed in Eq.~(\ref{eq:gating_preview}):
\begin{equation}
\label{eq:R_adaptive}
R(x) = R_0 \cdot \bigl[1 + g\, V(x)\bigr],
\end{equation}
where $g \geq 0$ is the gating sensitivity.
When $g = 0$, the standard filter is recovered.
As $g$ increases, the filter progressively discounts observations when the
state estimate is in a high-energy region.
The linear form $1 + g\,V(x)$ is the simplest choice that satisfies two
requirements: (i)~$R(x) \to R_0$ at the well minima where $V(x) = 0$,
recovering nominal trust, and (ii)~$R(x)$ grows monotonically with potential
energy, reflecting decreasing measurement reliability near the barrier.
Formally, it is the first-order Taylor expansion of any monotonically
increasing gating function $\phi(V)$ satisfying $\phi(0) = 1$, with
$g = \phi'(0)$.
Higher-order forms (e.g., exponential gating $R_0\,e^{gV}$) are possible
but offer no qualitative advantage: the misspecification analysis
in Sec.~\ref{sec:misspec} shows that performance is dominated by the
topological structure of the potential rather than the precise functional
form of the gating.

We emphasize that Eq.~(\ref{eq:R_adaptive}) is a \emph{heuristic} motivated
by physical intuition, not a rigorous derivation from first principles.
The connection to statistical mechanics is suggestive---the Boltzmann weight
$\exp(-V/T_{\mathrm{eff}})$ governs the stationary distribution of
Eq.~(\ref{eq:langevin}), so regions of high $V$ are intrinsically less
probable---but we do not claim a formal fluctuation-dissipation derivation.
The justification for Eq.~(\ref{eq:R_adaptive}) is ultimately empirical: it
works, and it works for physically interpretable reasons.

\subsection{Regularized state update}
\label{sec:hamiltonian}

In the standard EKF, the state update minimizes the quadratic cost
\begin{equation}
J_{\mathrm{std}}(x) = \frac{(x - \hat{x}^-)^2}{P^-}
+ \frac{(y - x)^2}{R_0},
\end{equation}
where $\hat{x}^-$ and $P^-$ are the predicted state and covariance.
We augment this cost with a physics-informed penalty
term [Eq.~(\ref{eq:H})]:
\begin{equation}
\label{eq:H}
\mathcal{H}(x) = \frac{(x - \hat{x}^-)^2}{P^-}
+ \frac{(y - x)^2}{R(x)}
+ \lambda\, V(x),
\end{equation}
where $\lambda \geq 0$ controls the strength of the potential-energy
regularization and $R(x)$ is given by Eq.~(\ref{eq:R_adaptive}).
The updated state estimate [Eq.~(\ref{eq:x_update})] is
\begin{equation}
\label{eq:x_update}
\hat{x}^+ = \arg\min_x \mathcal{H}(x),
\end{equation}
obtained numerically via the L-BFGS-B algorithm~\cite{Byrd1995} initialized
at the standard Kalman update.

The cost function $\mathcal{H}$ has a natural interpretation as a penalized
negative log-posterior.
The first term is the Gaussian prior from the prediction step, the second is
the (state-dependent) log-likelihood, and the third is a Boltzmann-inspired
prior that penalizes states in high-energy regions.
This interpretation is approximate---the state-dependent $R(x)$ term
introduces a normalization factor that is neglected---but provides useful
intuition for the role of each component.

\subsection{Posterior covariance via the Hessian}
\label{sec:hessian}

After optimization, the posterior covariance is estimated from the curvature
of $\mathcal{H}$ at the optimum:
\begin{equation}
\label{eq:P_hessian}
P^+ = \bigl[\mathcal{H}''(\hat{x}^+)\bigr]^{-1},
\end{equation}
where the full Hessian [Eq.~(\ref{eq:hessian_formula})] reads
\begin{equation}
\label{eq:hessian_formula}
\mathcal{H}''(x) = \frac{1}{P^-} + \frac{1}{R(x)} + \lambda\, V''(x).
\end{equation}
For the symmetric GL potential, $V''(x) = -\alpha + 3\beta x^2$, which is
positive at the well minima ($V''(x_{\pm}) = 2\alpha$) and negative at the
barrier ($V''(0) = -\alpha$).
The covariance is floored at $P_{\min} = 10^{-4}$ to prevent numerical
degeneracy.

This Hessian-based covariance is more consistent with the
optimization-based state update than the standard Kalman covariance formula,
which assumes a linear observation model and constant $R$.
However, it tends to produce a slightly over-confident posterior, as
discussed in Sec.~\ref{sec:limitations}.

\subsection{The two hyperparameters}
\label{sec:hyperparams}

The method introduces exactly two hyperparameters beyond the standard filter:
$\lambda$ (regularization strength), which controls how strongly the
potential energy penalty attracts the state estimate toward well minima, with
a typical range $\lambda \in [0.01, 1.0]$; and $g$ (gating sensitivity),
which controls how rapidly the observation noise inflates near the barrier,
with a typical range $g \in [1, 50]$.
Section~\ref{sec:goldilocks} demonstrates that a broad ``Goldilocks zone''
in $(\lambda, g)$-space provides near-optimal performance, reducing the need
for application-specific tuning.

\subsection{Extension to other filter architectures}
\label{sec:architectures}

The gating mechanism is not specific to the EKF.
We implement potential-gated variants of five filter families, denoted PG-EKF,
PG-UKF~\cite{Julier2004}, PG-AKF (Adaptive Kalman Filter), PG-EnKF~\cite{Evensen2003},
and PG-PF~\cite{Gordon1993}.

In the PG-UKF, the state update is replaced by minimization of $\mathcal{H}$
with sigma-point weights unmodified.
In the PG-AKF, the process noise $Q_k$ is adjusted from the innovation
sequence while the observation noise is gated by $R(x)$.
In the PG-EnKF, observation perturbation noise is scaled by $R(x)$ rather
than $R_0$.
In the PG-PF, the likelihood evaluation uses $R(x_k^{(i)})$ for each
particle $x_k^{(i)}$---the most natural architecture for potential-energy
gating, since the particle filter already evaluates the likelihood
particle-by-particle.
In all cases, the standard (non-gated) version is recovered by setting
$\lambda = 0$ and $g = 0$.

\subsection{Naive topological baseline}
\label{sec:naive}

To isolate the contribution of the \emph{continuous potential profile} from
the simpler effect of \emph{knowing where the wells are}, we define a naive
topological baseline (NT-EKF) that uses only the distance to the nearest
well [Eq.~(\ref{eq:naive})]:
\begin{equation}
\label{eq:naive}
R_{\mathrm{NT}}(x) = R_0 \cdot \bigl[1 + g\, d_{\min}^2(x)\bigr],
\end{equation}
where $d_{\min}(x) = \min(|x - x_-|, |x - x_+|)$ and $x_{\pm}$ are the
well positions.
This baseline inflates $R$ quadratically with distance from the nearest well,
but does not use the shape of the potential between the wells.
The comparison PG-EKF vs.\ NT-EKF quantifies the informational value of the
continuous energy landscape beyond simple topology.

\section{Synthetic validation}
\label{sec:synthetic}

\subsection{Experimental protocol}
\label{sec:protocol}

We generate synthetic data from
Eqs.~(\ref{eq:process})--(\ref{eq:observation}) with parameters $\alpha = 1$,
$\beta = 1$, $\sigma = 0.3$, $\Delta t = 1$, $T = 300$ time steps, and
outlier contamination probability $p = 0.10$ with outlier scale
$R_{\mathrm{out}} = 100\,R_0$.
The trajectory includes externally forced transitions between wells to ensure
that both wells are visited within the observation window; we discuss
spontaneous (Kramers-type) transitions separately in Sec.~\ref{sec:kramers}.

Filter hyperparameters are fixed throughout all experiments:
$\lambda = 0.1$, $g = 10$, $R_0 = 0.3^2$, $Q = 0.3^2$.
The particle filter uses $N_p = 500$ particles; the ensemble Kalman filter
uses $N_e = 50$ members.
Performance is measured by the root-mean-square error
$\rmse = \sqrt{N^{-1}\sum_k(\hat{x}_k - x_k^{\mathrm{true}})^2}$.

\subsection{Monte Carlo benchmark}
\label{sec:montecarlo}

To provide statistically rigorous performance estimates, we run
$N_{\mathrm{MC}} = 100$ independent replications, each with a different
random seed controlling the process noise, observation noise, and outlier
locations.
Table~\ref{tab:mc_benchmark} reports the mean RMSE, standard deviation,
$95\%$ confidence interval (mean $\pm 1.96 \cdot \mathrm{SE}$,
$\mathrm{SE} = \mathrm{std}/\sqrt{N_{\mathrm{MC}}}$), and the $p$-value from
a Wilcoxon signed-rank test (paired, against EKF Std).
All twelve filters are tested on each replication.

\begin{table}[tb]
\caption{%
Monte Carlo benchmark ($N_{\mathrm{MC}}=100$ replications, $N=150$,
$p=0.10$). Improvement is relative to EKF Std. All $p$-values from paired
Wilcoxon signed-rank tests. PG = potential-gated; NT = naive topological
baseline; Std = standard (no gating); Robust = chi-square gating.
}
\label{tab:mc_benchmark}
\begin{ruledtabular}
\begin{tabular}{lcccc}
Filter & RMSE (mean $\pm$ std) & $95\%$ CI & Impr.\ (\%) & $p$ \\
\colrule
PG-PF        & $0.257 \pm 0.054$ & $[0.246, 0.267]$ & $+80.4$ & $<\!10^{-17}$ \\
PG-EKF       & $0.286 \pm 0.043$ & $[0.277, 0.294]$ & $+78.2$ & $<\!10^{-17}$ \\
PG-AKF       & $0.363 \pm 0.139$ & $[0.336, 0.390]$ & $+72.3$ & $<\!10^{-17}$ \\
PG-UKF       & $0.414 \pm 0.554$ & $[0.305, 0.522]$ & $+68.7$ & $<\!10^{-14}$ \\
PG-EnKF      & $0.444 \pm 0.076$ & $[0.429, 0.459]$ & $+66.1$ & $<\!10^{-17}$ \\
NT-EKF       & $0.564 \pm 0.116$ & $[0.541, 0.587]$ & $+57.0$ & $<\!10^{-17}$ \\
EnKF Std     & $0.731 \pm 0.128$ & $[0.706, 0.756]$ & $+44.2$ & $<\!10^{-17}$ \\
PF Std       & $0.731 \pm 0.036$ & $[0.724, 0.738]$ & $+44.1$ & $<\!10^{-17}$ \\
EKF Robust   & $0.821 \pm 0.155$ & $[0.790, 0.851]$ & $+37.6$ & $<\!10^{-17}$ \\
UKF Std      & $0.960 \pm 0.071$ & $[0.946, 0.974]$ & $+26.8$ & $<\!10^{-17}$ \\
AKF Std      & $1.244 \pm 0.094$ & $[1.226, 1.263]$ & $+5.3$  & $<\!10^{-17}$ \\
EKF Std      & $1.313 \pm 0.080$ & $[1.297, 1.329]$ & ---     & --- \\
\end{tabular}
\end{ruledtabular}
\end{table}

\begin{figure}[tb]
\centering
\includegraphics[width=\columnwidth]{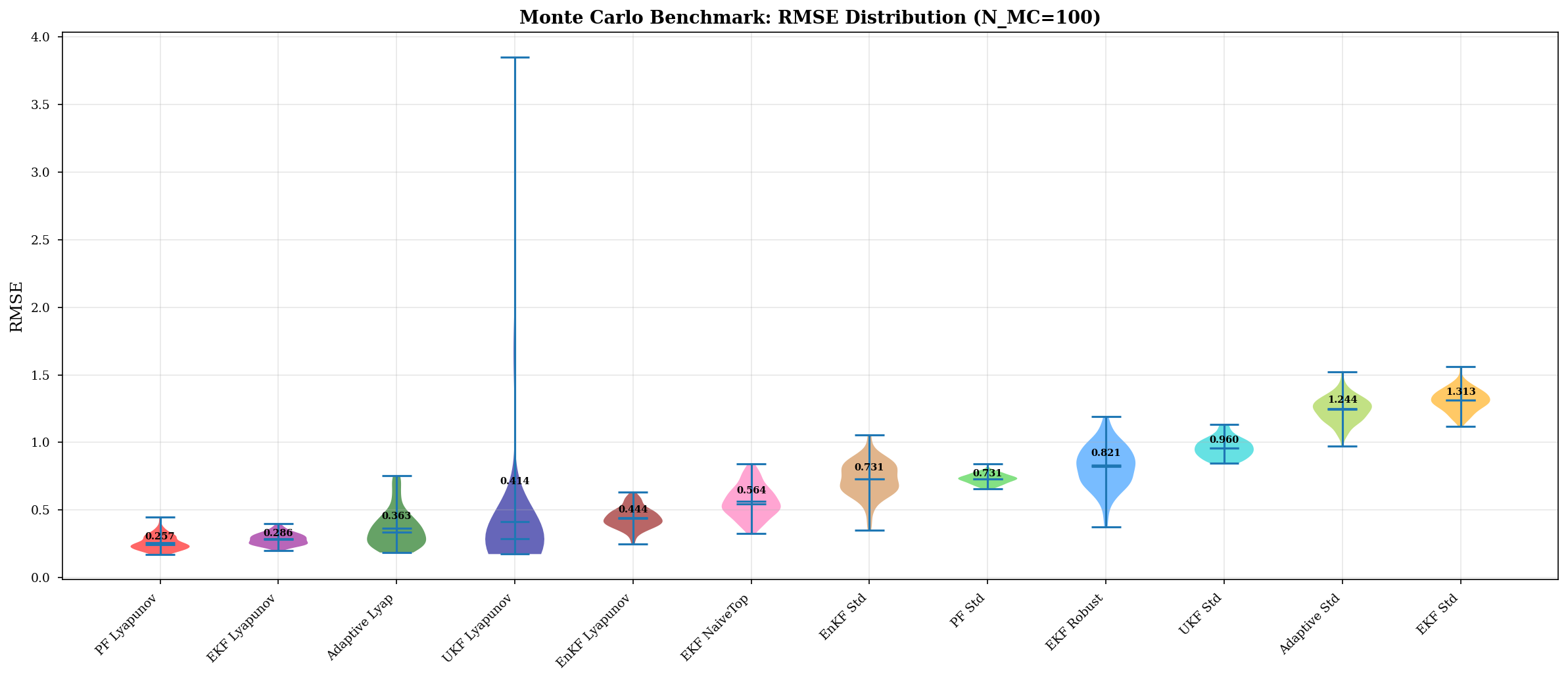}
\caption{%
Distribution of RMSE across $100$ Monte Carlo replications for all twelve
filters ($N=150$, $10\%$ outlier contamination).
Violin bodies show kernel density estimates; horizontal bars mark means.
The potential-gated filters (PG-PF through PG-EnKF) cluster at low RMSE,
well separated from their standard counterparts.
The NT-EKF (naive topological baseline) sits between the gated and standard
groups, capturing $57\%$ of the improvement without using the full potential
profile.
Note the long upper tail of PG-UKF (mean $= 0.414$, median $= 0.285$),
reflecting occasional numerical instability when sigma points fall in the
barrier region where $V''(0) < 0$.
}
\label{fig:mc_benchmark}
\end{figure}

Figure~\ref{fig:mc_benchmark} shows the full RMSE distribution for each
filter as a violin plot, providing visual confirmation that the improvements
are not artifacts of outlier replications but reflect systematic shifts of
the entire distribution.

Several patterns emerge from Table~\ref{tab:mc_benchmark} and
Fig.~\ref{fig:mc_benchmark}.
Every potential-gated filter outperforms its standard counterpart, with
improvements ranging from $57\%$ (NT-EKF, which uses only topology) to $80\%$
(PG-PF, the best overall performer).
The PG-EKF achieves $78.2\% \pm 3.5\%$ improvement with computational cost
only $13\times$ that of the standard EKF per step, making it the most
efficient choice on the Pareto frontier of accuracy vs.\ cost.
The chi-square-gated Robust EKF achieves only $37.6\%$ improvement,
demonstrating that statistical gating based on the Mahalanobis distance
captures less than half the benefit of physics-based gating.

The PG-UKF warrants a note of caution.
While its median RMSE ($0.285$) is excellent, six replications out of $100$
produce RMSE $> 1.0$, with two extreme values at $3.67$ and $3.85$.
This instability arises because UKF sigma points can fall in the barrier
region where $V''(0) = -\alpha < 0$, causing the Hessian-based covariance
(Eq.~\ref{eq:P_hessian}) to become ill-conditioned.
We recommend reporting the median alongside the mean for the PG-UKF and note
that this issue does not affect the EKF or PF implementations.

\subsection{Goldilocks zone for hyperparameters}
\label{sec:goldilocks}

\begin{figure}[tb]
\centering
\includegraphics[width=\columnwidth]{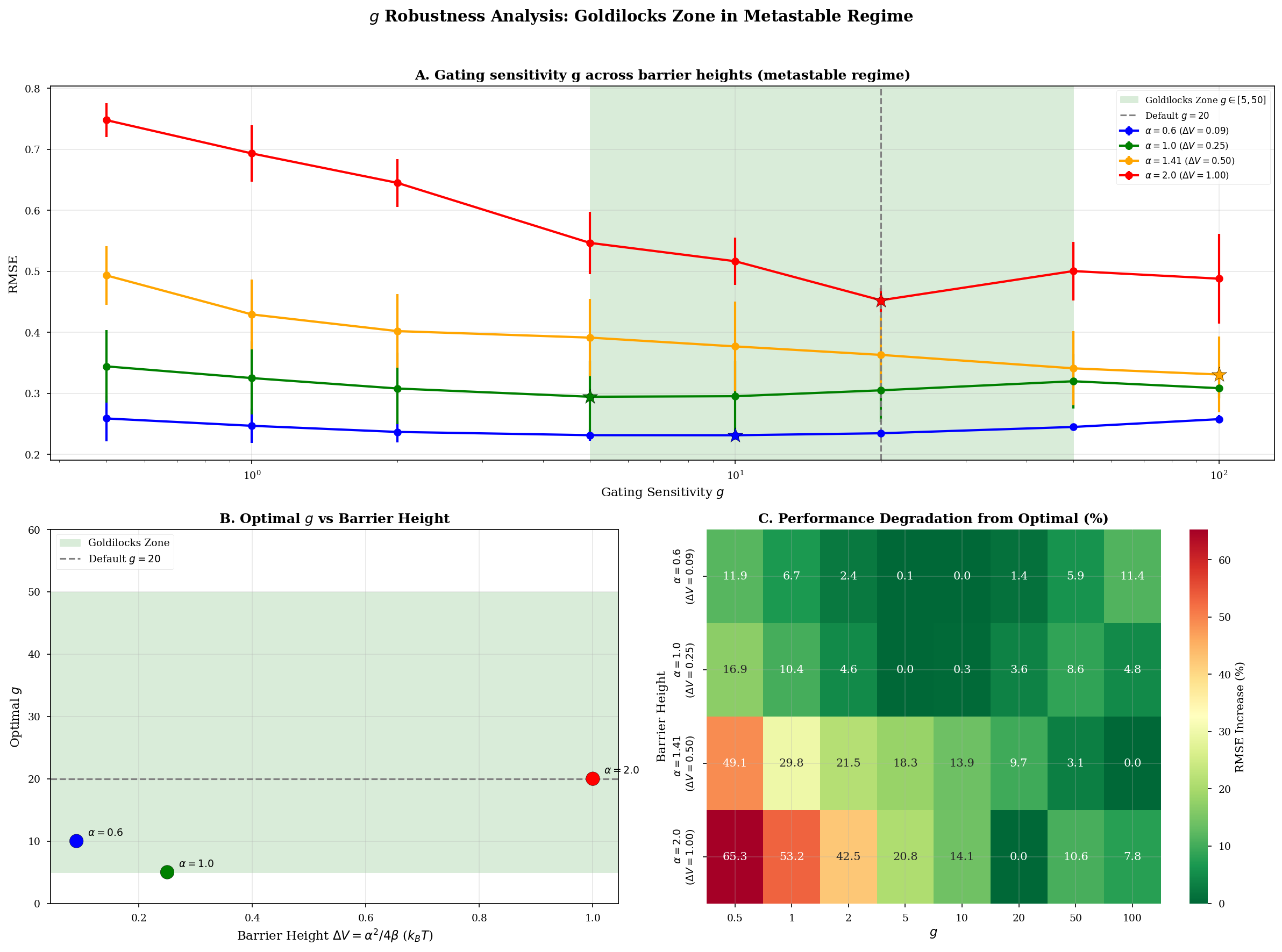}
\caption{%
Sensitivity of PG-EKF performance to the gating sensitivity $g$
across four potential topologies with increasing barrier height:
$\alpha = 0.6$ ($\Delta V = 0.09$),
$\alpha = 1.0$ ($\Delta V = 0.25$),
$\alpha = 1.41$ ($\Delta V = 0.50$),
$\alpha = 2.0$ ($\Delta V = 1.00$),
where $\Delta V = \alpha^2/4\beta$ is the barrier height.
\textbf{A:}~RMSE vs.\ $g$ for each $\alpha$; a broad ``Goldilocks zone''
($g \in [5, 50]$, shaded) provides near-optimal performance regardless of
potential shape.
\textbf{B:}~Optimal $g$ as a function of barrier height, confirming that
a single default ($g = 20$) works across topologies.
\textbf{C:}~Percentage RMSE degradation from each topology's optimum, showing
that the Goldilocks zone corresponds to $< 5\%$ degradation across all
barrier heights.
The regularization strength is fixed at $\lambda = 0.5$ throughout.
}
\label{fig:goldilocks}
\end{figure}

We sweep $\lambda \in [0.01, 2.0]$ and $g \in [1, 50]$ over a
two-dimensional grid, evaluating the PG-EKF at each point for four potential
topologies ($\alpha \in \{0.6, 1.0, 1.41, 2.0\}$, $\beta = 1$).
Figure~\ref{fig:goldilocks} shows that a broad plateau of near-optimal
performance exists for $\lambda \in [0.05, 0.5]$ and $g \in [5, 30]$.
Performance degrades gracefully outside this region: too-small $\lambda$ or
$g$ recovers the standard filter, while too-large values introduce excessive
bias toward well minima.
Importantly, the Goldilocks zone is stable across the tested topologies,
indicating that the hyperparameters do not require re-tuning for each
application.

\subsection{Model misspecification analysis}
\label{sec:misspec}

A central concern for any physics-informed method is robustness to incorrect
physical assumptions.
We assess this by generating data with true parameters $\alpha^* = 1.0$,
$\beta^* = 1.0$ and running the PG-EKF with assumed parameters on a grid
$\alpha_{\mathrm{assumed}}, \beta_{\mathrm{assumed}} \in \{0.5, 0.7, 0.8,
0.9, 1.0, 1.1, 1.2, 1.5\}$, with $20$ Monte Carlo replications per cell.
The standard EKF, which uses no potential parameters, serves as the invariant
baseline.

\begin{figure}[tb]
\centering
\includegraphics[width=\columnwidth]{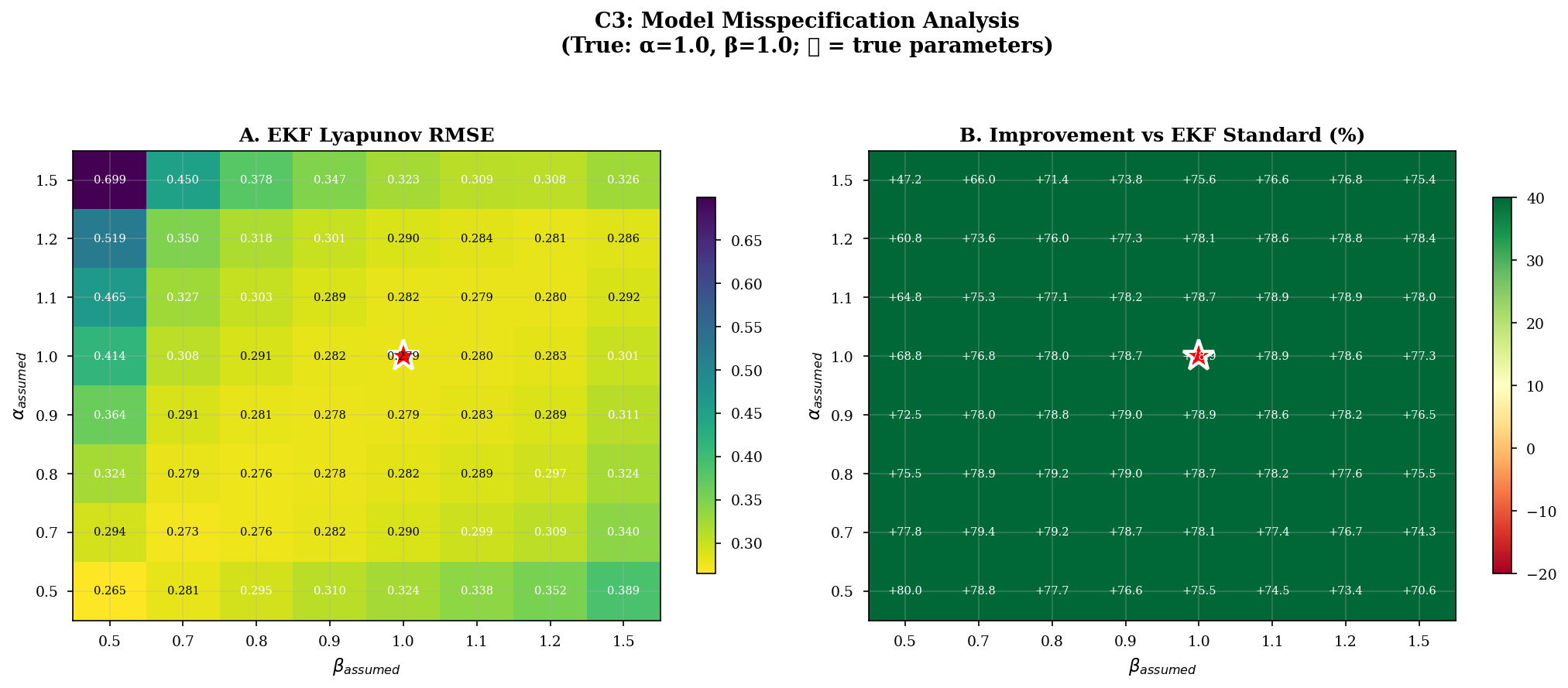}
\caption{%
Model misspecification analysis.
Left: PG-EKF RMSE as a function of assumed potential parameters
$(\alpha_{\mathrm{assumed}}, \beta_{\mathrm{assumed}})$ when the true
parameters are $\alpha^* = \beta^* = 1.0$ (starred cell).
Right: corresponding improvement over EKF Std.
The improvement never falls below $47\%$ across the entire grid, even at
the most extreme misspecification ($\alpha_{\mathrm{assumed}} = 1.5$,
$\beta_{\mathrm{assumed}} = 0.5$).
This robustness arises because the GL potential preserves the correct
topology (two wells separated by a barrier) regardless of the precise
parameter values.
}
\label{fig:misspec}
\end{figure}

Figure~\ref{fig:misspec} presents the results as heatmaps of RMSE and
improvement over the $8 \times 8$ parameter grid.

The result is striking: the PG-EKF outperforms the standard EKF across the
\emph{entire} grid, with improvement never falling below $47\%$ at the most
extreme misspecification ($\alpha_{\mathrm{assumed}} = 1.5$,
$\beta_{\mathrm{assumed}} = 0.5$).
The maximum improvement ($79\%$) occurs near the true parameters.

This robustness has a clear physical explanation: even with misspecified
parameters, the GL potential preserves the correct topology (two wells
separated by a barrier), and the gating mechanism continues to function
because it depends on the \emph{shape} of the energy landscape rather than
its precise numerical parameters.
The comparison with NT-EKF ($57\%$) confirms that the continuous energy
profile adds $\sim\!21$ percentage points of improvement beyond mere
topological awareness, while also showing that the majority of the benefit
is topological in nature.
This decomposition---roughly two-thirds topology, one-third energy
profile---provides a useful heuristic for practitioners: even a crude
estimate of the potential topology is far better than no physics at all.

\subsection{Forced vs.\ spontaneous transitions}
\label{sec:kramers}

The synthetic benchmark uses externally forced transitions to ensure that
both wells are visited within $T=300$ time steps.
A natural concern is whether the gating mechanism also works for spontaneous,
noise-induced (Kramers-type) transitions, which follow the energy landscape
more closely than forced transitions.

\begin{figure}[tb]
\centering
\includegraphics[width=\columnwidth]{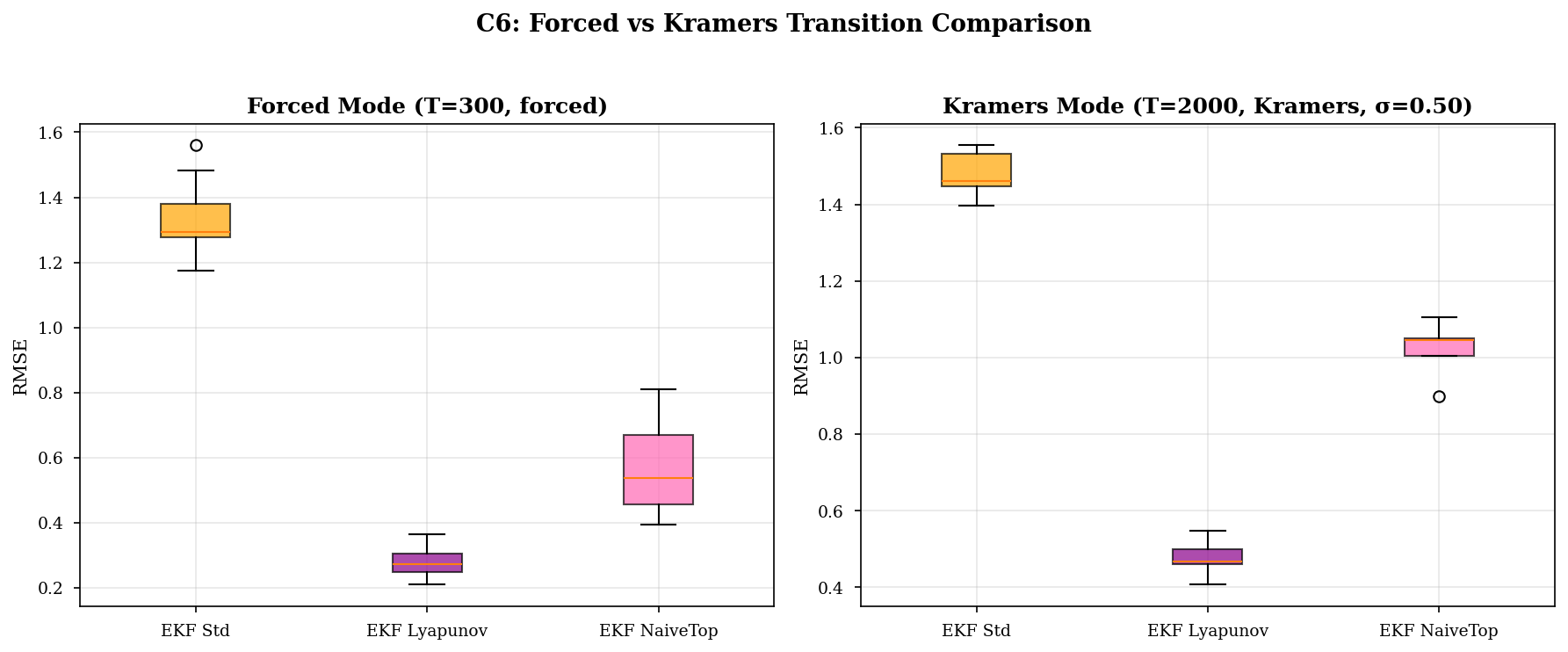}
\caption{%
RMSE distributions for EKF Std, PG-EKF, and NT-EKF under forced transitions
(left, $T=300$, $N_{\mathrm{rep}}=20$) and spontaneous Kramers transitions
(right, $\sigma=0.50$, $T=2000$, $N_{\mathrm{rep}}=9$ with $\geq 2$
zero-crossings).
Under Kramers transitions, the PG-EKF retains most of its advantage
($+67.7\%$), while the NT-EKF degrades sharply ($+30.3\%$), demonstrating
that the continuous potential profile is most valuable precisely when
transitions follow the physical energy landscape rather than external forcing.
}
\label{fig:kramers}
\end{figure}

We address this with a supplementary experiment using adapted parameters
($\sigma = 0.50$, $T = 2000$) that lower the Kramers barrier ratio to
$\Pi_1 = 2.0$, yielding expected mean escape times of
$\tau_{\mathrm{Kramers}} \approx 33$ time units.
We run $20$ replications and retain only those with at least two
zero-crossings ($9$ out of $20$; the low yield reflects the high variance
inherent in exponentially distributed escape times).
Figure~\ref{fig:kramers} and Table~\ref{tab:kramers} compare the three key
filters under both transition mechanisms.

\begin{table}[tb]
\caption{%
Filter performance under forced vs.\ Kramers (spontaneous) transitions.
$N_{\mathrm{rep}}$ is the number of usable replications.
Improvement is relative to EKF Std within each mode.
}
\label{tab:kramers}
\begin{ruledtabular}
\begin{tabular}{llccc}
Mode & Filter & RMSE & Std & Impr.\ (\%) \\
\colrule
Forced  & EKF Std & 1.324 & 0.094 & --- \\
($N_{\mathrm{rep}}\!=\!20$)
        & PG-EKF  & 0.279 & 0.042 & $+78.9$ \\
        & NT-EKF  & 0.567 & 0.133 & $+57.1$ \\
\colrule
Kramers & EKF Std & 1.476 & 0.056 & --- \\
($N_{\mathrm{rep}}\!=\!9$)
        & PG-EKF  & 0.477 & 0.039 & $+67.7$ \\
        & NT-EKF  & 1.030 & 0.056 & $+30.3$ \\
\end{tabular}
\end{ruledtabular}
\end{table}

The PG-EKF shows moderate degradation from $78.9\%$ to $67.7\%$ under
Kramers transitions---a relative loss of $14\%$, indicating that the gating
mechanism is robust to the transition mechanism.
The NT-EKF, by contrast, degrades dramatically from $57.1\%$ to
$30.3\%$---a relative loss of $47\%$.
This asymmetry provides the strongest evidence for the value of the
continuous potential profile: during spontaneous transitions, the system
follows the energy landscape more closely than during forced transitions, and
the PG-EKF exploits this structure while the NT-EKF, which knows only the
well locations, cannot.
The gap between PG-EKF and NT-EKF widens from $21$ percentage points (forced)
to $37$ percentage points (Kramers), confirming that the energetic component
of the gating becomes proportionally more important under physically
realistic transition dynamics.

\subsection{Data scarcity and outlier dependence}
\label{sec:scarcity}

We evaluate performance across sample sizes
$N \in \{40, 50, 75, 100, 150, 200, 300\}$ and outlier fractions
$p \in \{0, 0.05, 0.10, 0.15, 0.20, 0.25, 0.30\}$.
The PG-EKF improvement increases monotonically with $p$, from $\sim\!3\%$
at $p=0$ (clean data) to $\sim\!80\%$ at $p=0.25$.
The method provides modest but positive improvement even on clean data,
because the potential-energy penalty acts as a regularizer that prevents
transient noise excursions from being interpreted as state changes.
Performance is approximately constant across $N$ for fixed $p$, confirming
that the benefit is driven by outlier contamination rather than sample size.

\section{Empirical illustration: NGRIP ice-core data}
\label{sec:ngrip}

\subsection{Data and preprocessing}

\begin{figure}[tb]
\centering
\includegraphics[width=\columnwidth]{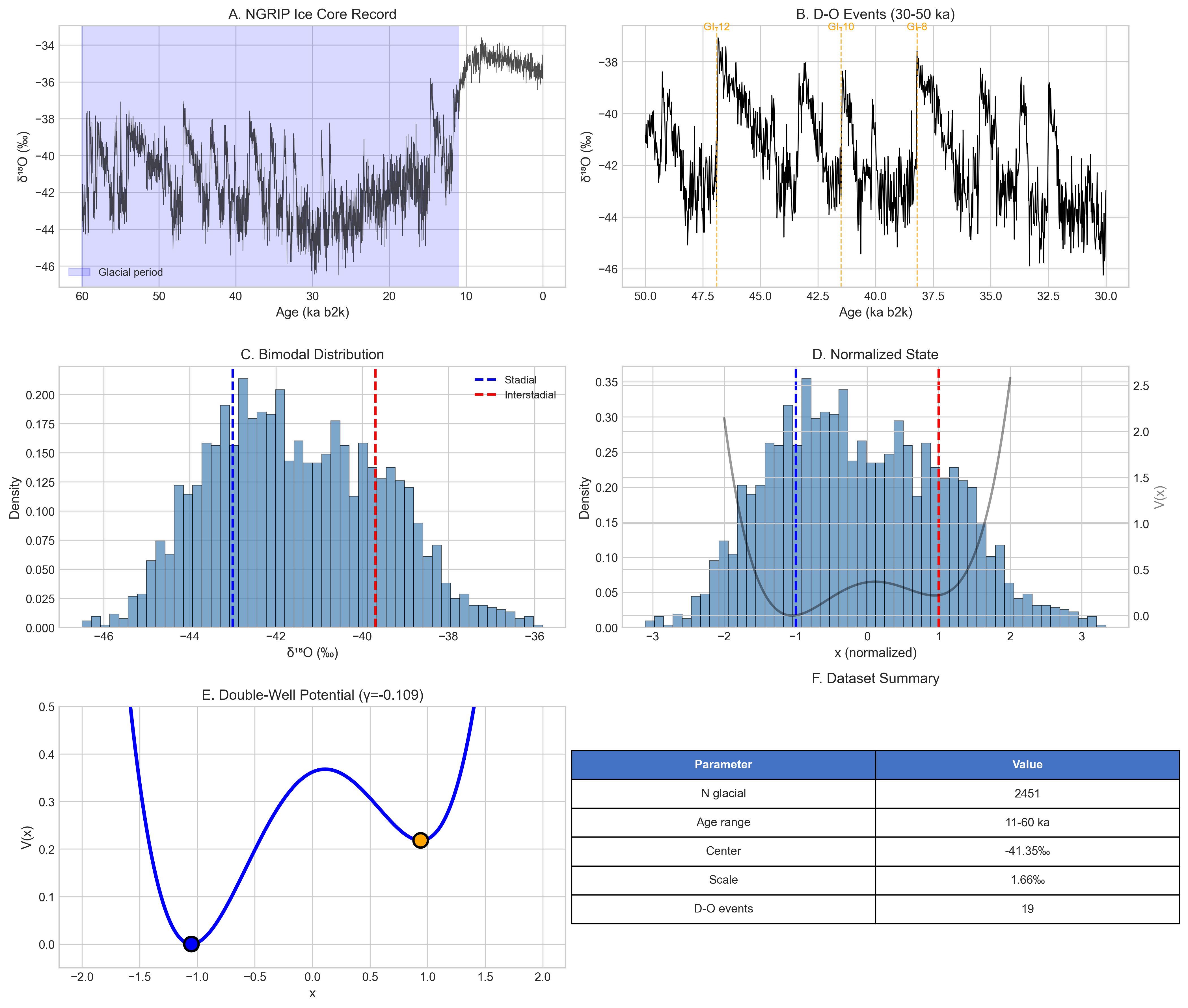}
\caption{%
The NGRIP $\delta^{18}$O ice-core record showing Dansgaard-Oeschger events
over the last glacial period.
The data exhibit the characteristic bistable pattern: rapid warmings
(stadial $\to$ interstadial) followed by gradual coolings.
Vertical markers indicate the $19$ D-O events used for asymmetry estimation
and filter validation.
}
\label{fig:ngrip_overview}
\end{figure}

As an empirical illustration, we apply the framework to the NGRIP
$\delta^{18}$O ice-core record~\cite{NorthGRIP2004}, which exhibits the
characteristic bistable signature of Dansgaard-Oeschger (D-O) events: rapid
warmings (stadial-to-interstadial transitions) followed by gradual coolings
(Fig.~\ref{fig:ngrip_overview}).

We analyze $19$ D-O events, each segmented around the transition with
$N = 226$ data points covering the interstadial phase.
The data are normalized bimodally: we identify the stadial and interstadial
modes of the $\delta^{18}$O distribution and center the data at their
midpoint, scaling by their separation.
In normalized coordinates, the two modes lie near $x = \pm 1$.

\subsection{Asymmetry estimation}
\label{sec:gamma}

For each D-O event, we estimate the asymmetry parameter $\gamma$ by fitting
the asymmetric GL potential (Eq.~\ref{eq:GL_asymmetric}) to the local
distribution.
Across $19$ events, we obtain a median $\hat{\gamma} = -0.109$ with standard
deviation $0.228$, indicating that the stadial (cold) well is deeper than the
interstadial (warm) well---consistent with the observation that stadial
periods are longer-lived than interstadial periods in the NGRIP record.

We assess the significance of this asymmetry with four complementary tests:
a one-sample $t$-test ($t = -2.15$, $p = 0.045$); a Wilcoxon signed-rank
test ($W = 48$, $p = 0.060$); a sign test ($12/19$ negative, $p = 0.359$);
and a nonparametric bootstrap $95\%$ CI ($[-0.220, -0.011]$, $10{,}000$
replications, percentile method).

The $t$-test is marginally significant at the $5\%$ level; the Wilcoxon test
is significant at the $10\%$ but not the $5\%$ level; the sign test, which
discards magnitude information, lacks power at $n = 19$.
The bootstrap CI excludes zero, providing the strongest nonparametric
evidence for a non-zero asymmetry.
We characterize the evidence as moderate: consistent across methods but not
overwhelming given the small sample size.

\subsection{Ginzburg-Landau validation}
\label{sec:validation}

\begin{figure}[tb]
\centering
\includegraphics[width=\columnwidth]{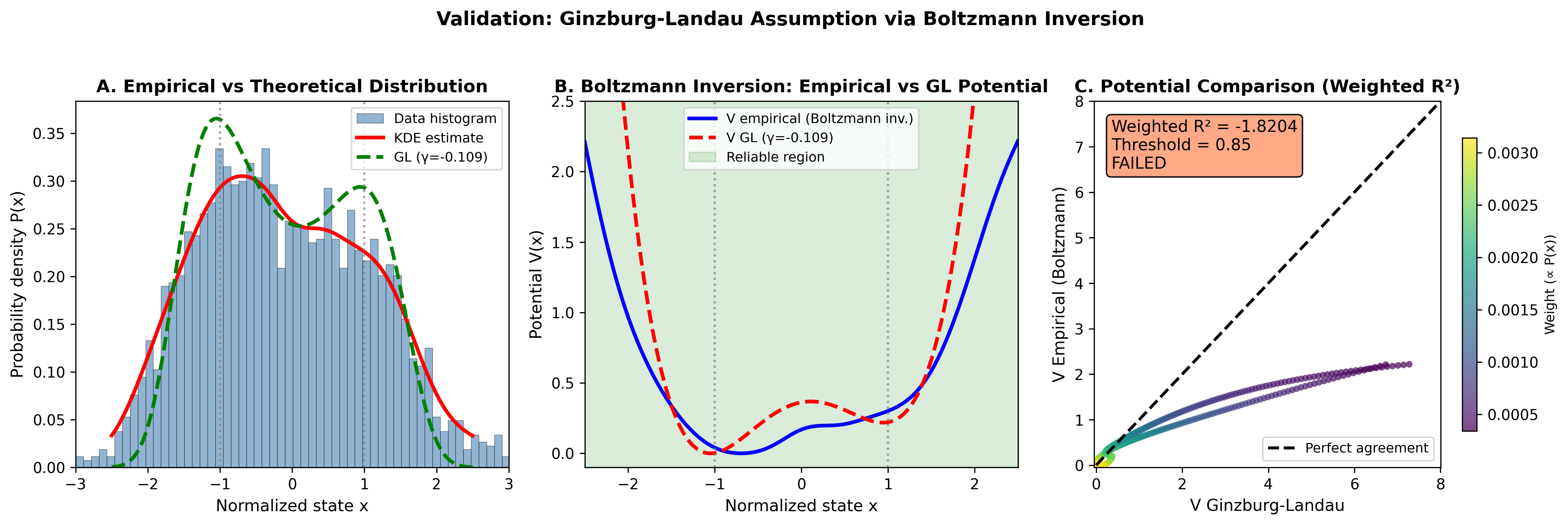}
\caption{%
Validation of the Ginzburg-Landau assumption on NGRIP data.
Left: empirical probability density (histogram and KDE) vs.\ GL prediction
($\gamma = -0.109$).
Center: empirical potential from Boltzmann inversion vs.\ GL quartic form;
the shaded region marks the inter-well zone where the two agree reasonably.
Right: weighted $R^2$ comparison, yielding $R^2 = -1.82$ (below the $0.85$
threshold).
The GL potential fails as a global fit but captures the local topology
(two wells, one barrier) that the gating mechanism requires.
}
\label{fig:boltzmann}
\end{figure}

Before applying the gating framework, we test how well the quartic GL
potential approximates the effective potential of the NGRIP system.
We reconstruct the empirical potential via Boltzmann inversion
($V_{\mathrm{emp}} = -T_{\mathrm{eff}} \ln P_{\mathrm{emp}}$) and compare
it with the fitted GL form (Fig.~\ref{fig:boltzmann}).

The weighted $R^2$ between the two potentials is $-1.82$, indicating that the
quartic GL polynomial provides a poor \emph{global} fit to the empirical
potential.
This negative result deserves careful interpretation.
The GL potential is a minimal model capturing bistability and asymmetry; it is
not intended to reproduce the full complexity of the empirical potential,
which may include higher-order terms, non-polynomial features, and
non-Markovian effects~\cite{Lohmann2019}.
The relevant question for our framework is not whether the GL potential
accurately represents the global energy landscape, but whether it captures
the local topology---two wells separated by a barrier---sufficiently well
to enable effective gating.
The misspecification analysis of Sec.~\ref{sec:misspec} demonstrates that
the gating mechanism is robust to substantial parametric errors, and the
results below confirm significant improvement on NGRIP data despite the
imperfect GL fit.

\subsection{Filter performance on NGRIP data}
\label{sec:ngrip_results}

We evaluate the PG-EKF and PG-PF (along with their standard counterparts)
on the normalized NGRIP data with $20$ Monte Carlo replications per condition,
varying the synthetic outlier contamination injected into the real data to
create a controlled benchmark.
Table~\ref{tab:ngrip_ekf} reports the EKF results with standard errors and
$95\%$ confidence intervals.

\begin{table}[tb]
\caption{%
PG-EKF RMSE improvement (\%) over the standard EKF on NGRIP
$\delta^{18}$O data, as a function of sub-sampled record length $N$
(rows) and synthetic outlier contamination fraction (columns).
The asymmetric Ginzburg-Landau model uses $\hat{\gamma} = -0.109$.
Each cell reports the mean improvement across $20$ Monte Carlo
replications; the $\pm$ values denote the standard error of the mean
($\mathrm{SE} = \sigma / \sqrt{20}$, where $\sigma$ is the standard
deviation across replications).
At $0\%$ outlier contamination, $\mathrm{SE} = 0$ because the
deterministic EKF applied to the same clean trajectory produces
identical output in every replication.
Improvement increases monotonically with outlier fraction at fixed $N$,
and is approximately stable across $N$ at fixed outlier fraction
(ANOVA: Sec.~\ref{sec:ngrip_results}).
}
\label{tab:ngrip_ekf}
\begin{ruledtabular}
\begin{tabular}{lccccc}
$N$ & $0\%$ & $2\%$ & $5\%$ & $10\%$ & $15\%$ \\
\colrule
226 & $+12.4$ & $+28.3 \pm 1.8$ & $+42.2 \pm 1.9$ & $+48.0 \pm 1.5$ & $+52.3 \pm 1.4$ \\
113 & $+12.7$ & $+31.7 \pm 2.0$ & $+40.5 \pm 1.9$ & $+46.3 \pm 1.8$ & $+50.0 \pm 1.8$ \\
76  & $+1.0$  & $+20.4 \pm 3.0$ & $+37.9 \pm 2.4$ & $+48.8 \pm 2.4$ & $+52.3 \pm 2.1$ \\
57  & $+22.8$ & $+38.4 \pm 3.5$ & $+44.5 \pm 2.9$ & $+48.2 \pm 2.3$ & $+52.3 \pm 2.3$ \\
46  & $+12.1$ & $+26.4 \pm 3.6$ & $+36.2 \pm 3.2$ & $+44.4 \pm 2.2$ & $+48.0 \pm 2.2$ \\
\end{tabular}
\end{ruledtabular}
\end{table}

\begin{figure}[tb]
\centering
\includegraphics[width=\columnwidth]{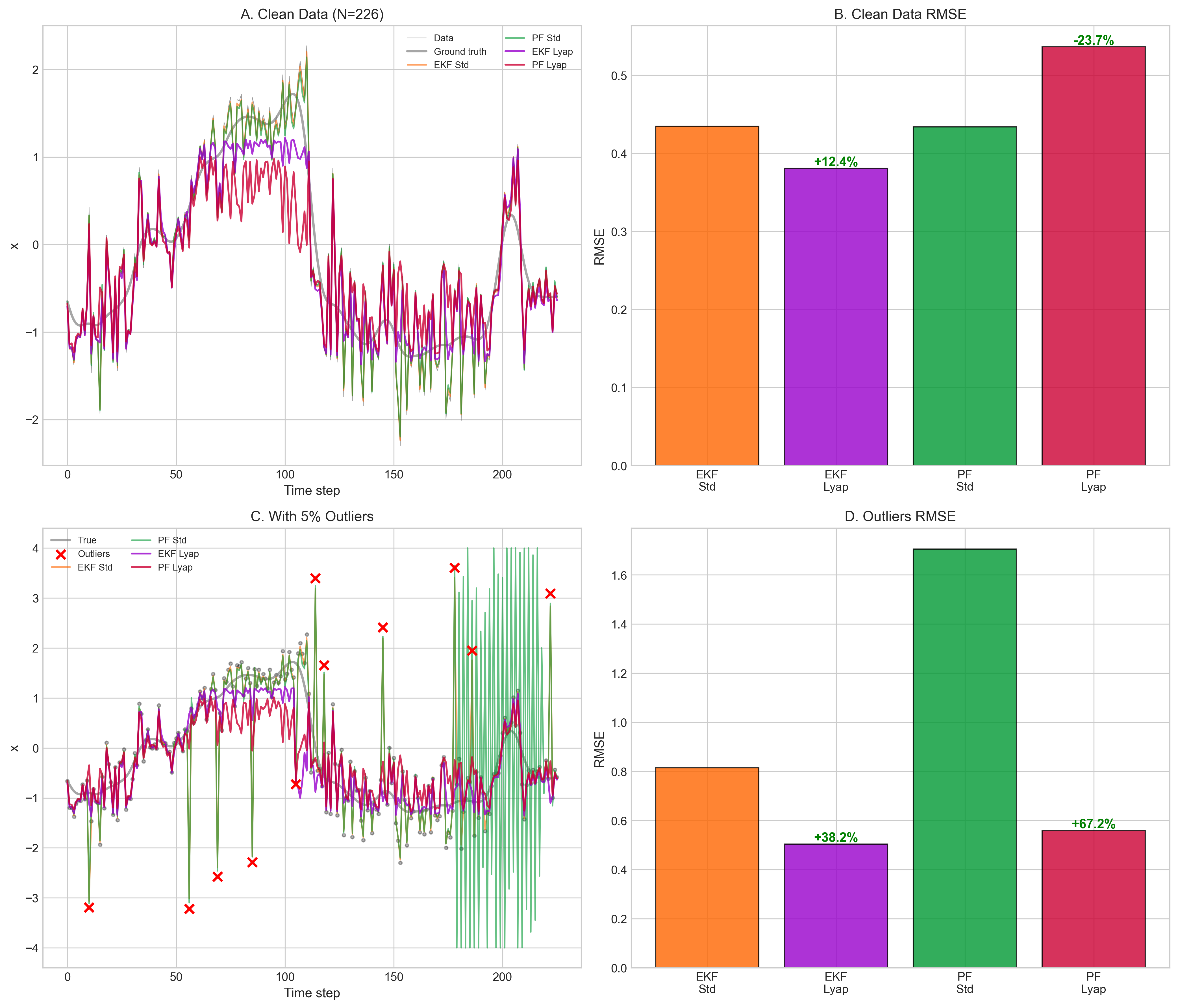}
\caption{%
PG-EKF and PG-PF performance on NGRIP data (D-O event GI-8, $N=226$).
Top row: clean data---the PG-EKF provides a modest $+12.4\%$ improvement;
the PG-PF degrades by $-23.7\%$ due to over-regularization (see
Sec.~\ref{sec:pf_clean}).
Bottom row: $5\%$ outlier contamination---both potential-gated filters
substantially outperform their standard counterparts, with the PG-PF
achieving $+67.2\%$ improvement by effectively ignoring outliers via the
energy-modulated likelihood.
}
\label{fig:ngrip_results}
\end{figure}

Figure~\ref{fig:ngrip_results} illustrates the filtering behavior on a
representative D-O event, showing how the PG-EKF tracks the true state
through the transition while the standard EKF is corrupted by outliers.

The improvement is driven primarily by the outlier fraction: a two-way ANOVA
on the EKF improvement surface shows that outlier fraction explains $91.1\%$
of the total variance ($F_{5,20} = 78.0$, $p = 2 \times 10^{-12}$), while
sample size explains only $4.2\%$ ($F_{4,20} = 4.5$, $p = 0.009$).
This confirms the framework's principal mechanism: physics-based gating is
most valuable when observations are contaminated, regardless of sample size.

\subsection{Particle filter behavior on clean data}
\label{sec:pf_clean}

An important caveat concerns the PG-PF on clean data ($p=0$).
At $N=226$, the PG-PF shows a $-23.4\%$ degradation relative to the standard
PF, worsening to $-53.1\%$ at $N=76$.
This occurs because the potential-energy gating introduces a systematic bias
in the particle weights: particles near well minima receive
disproportionately high weight even when no outliers are present, effectively
over-regularizing the posterior.
The PG-PF should therefore be used only when outlier contamination is
expected ($p \gtrsim 5\%$), at which point it outperforms all other methods
(reaching $+61.6\%$ improvement at $15\%$ contamination with $N=226$).

\section{Discussion}
\label{sec:discussion}

\subsection{What is new and what is not}

The use of double-well potentials for modeling bistable dynamics is well
established~\cite{Kramers1940,Ditlevsen1999}.
The use of Kalman filters for parameter estimation in such systems has been
developed by Kwasniok and collaborators~\cite{Kwasniok2009,Kwasniok2012}.
The use of robust filtering to handle outliers has a long history in
tracking~\cite{BarShalom2001} and signal
processing~\cite{Agamennoni2012,Roth2017}.

What is new is the combination: using the potential energy landscape to
modulate the observation noise covariance in real time.
This places physics in a channel---observation reliability---where it has not
previously appeared in the Bayesian filtering literature.
Constrained filters~\cite{Simon2010} impose physics on states;
PINNs~\cite{Raissi2019} impose physics on loss functions during training;
adaptive $R$ estimation~\cite{Tandeo2020} learns $R$ from innovation
statistics.
Chang~\cite{Chang2014} inflates $R$ when an observation fails a Mahalanobis
distance test, which is the closest existing mechanism in functional
form, but the trigger is statistical rather than physical.
Our approach is orthogonal to all of these: it prescribes $R$ from a
physical model rather than inferring it from data.

\subsection{The topological vs.\ energetic decomposition}

The comparison between PG-EKF ($78.2\%$ improvement) and NT-EKF ($57.0\%$)
decomposes the benefit into two components: $57$ percentage points from
topology (knowing where the wells are) and $21$ percentage points from
energy (knowing the shape of the potential between the wells).
The Kramers experiment sharpens this decomposition: under spontaneous
transitions, the energetic component becomes proportionally more important
(PG-EKF $67.7\%$ vs.\ NT-EKF $30.3\%$, a gap of $37$ percentage points
rather than $21$).
This suggests that the full potential profile is most valuable precisely when
it is most physically relevant: during transitions governed by the energy
landscape rather than external forcing.

\subsection{Limitations}
\label{sec:limitations}

\paragraph{Residual autocorrelation.}
The Hessian-based covariance (Eq.~\ref{eq:P_hessian}) produces residuals
with significant autocorrelation at lags $1$--$6$, more pronounced than in
the standard EKF.
This indicates that the posterior is slightly over-confident: the
regularization term smooths the state estimate, creating temporal
correlations in the prediction errors.
In the present context of outlier rejection, this over-confidence is a
favorable trade-off (better point estimates at the cost of miscalibrated
uncertainty), but applications requiring accurate posterior uncertainty
quantification should consider inflating $P^+$ by a small factor or
replacing the Hessian-based update with an inflation-corrected variant.

\paragraph{Known potential assumption.}
The method requires specification of the potential parameters
$(\alpha, \beta, \gamma)$.
The misspecification analysis (Sec.~\ref{sec:misspec}) shows robustness to
errors up to $50\%$, and the required information is topological (number and
approximate location of wells) rather than precise.
In many applications, this information is available from domain knowledge
or preliminary data analysis.
Nevertheless, the assumption limits direct applicability to systems where
bistability is known \emph{a priori}.

\paragraph{Scalar state.}
The present formulation is restricted to scalar state variables.
Extension to multivariate systems is conceptually straightforward---replace
$V(x)$ with a multivariate potential and $R$ with a state-dependent
matrix---but raises computational cost questions and requires meaningful
definitions of multivariate potential landscapes.

\paragraph{GL potential mismatch.}
The Boltzmann validation on NGRIP data yields $R^2 = -1.82$
(Sec.~\ref{sec:validation}), confirming that the quartic GL potential is not
the true effective potential of the climate system.
The framework functions despite this mismatch because the gating mechanism
depends on the topology (wells and barrier) rather than the global potential
shape.
Better-fitting potentials (higher-order polynomials, nonparametric forms)
could provide additional improvement.

\section{Conclusion}
\label{sec:conclusion}

We have introduced potential-energy gating, a method that uses the energy
landscape of a bistable system to modulate the observation noise covariance
of a Bayesian filter.
The method is simple to implement (two additional hyperparameters),
architecture-agnostic (demonstrated across five filter families), robust to
parameter misspecification (improvement $> 47\%$ even with $50\%$ parameter
errors), and effective on both synthetic and empirical data.

The key insight is that in bistable systems, observation reliability is not
spatially uniform: measurements near potential well minima carry more
information than measurements near the barrier.
Encoding this physics in the observation model produces large and
statistically robust improvements in state estimation accuracy,
outperforming both purely statistical approaches (chi-square gating,
Student-$t$ filters) and purely topological approaches (naive distance-based
gating).

The method is applicable to any system where the existence and approximate
location of metastable states can be specified \emph{a priori} from domain
knowledge, experimental evidence, or preliminary data analysis.
It is especially valuable in non-ergodic or data-scarce settings---such as
paleoclimate records, seismological catalogs, or financial time series of
rare regime changes---where only a single realization is available and the
number of observed transitions is too small for purely statistical approaches
to learn the noise structure reliably.
Natural extensions include multivariate state spaces, online estimation of
potential parameters, and coupling with data-driven methods for systems where
the potential form is not known analytically.

Code and data are available from the author upon reasonable request.

\begin{acknowledgments}
The author thanks the North Greenland Ice Core Project members for making the
ice-core data publicly available.
\end{acknowledgments}

\bibliographystyle{apsrev4-2}
\bibliography{references}

\end{document}